\pdfoutput=1

\documentclass[11pt]{article}

\usepackage{emnlp2021}

\usepackage{times}
\usepackage{latexsym}
\usepackage{graphicx}
\usepackage{tabularx}
\usepackage{multirow}
\usepackage{multicol}
\usepackage{adjustbox}
\usepackage{varwidth}
\usepackage{xspace}
\usepackage{microtype}
\usepackage{floatrow}
\usepackage{booktabs}

\usepackage[T1]{fontenc}

\usepackage[utf8]{inputenc}

\usepackage{microtype}

\usepackage{lingmacros}

\title{A Comparative Study on \\Collecting High-Quality Implicit Reasonings at a Large-scale}

\author{
Keshav Singh${}^{\,\ddagger}$
\quad\hspace{-5pt}
Paul Reisert${}^{\,\dagger,\ddagger}$
\quad\hspace{-5pt}
Naoya Inoue${}^{\,\star,\dagger}$
\quad\hspace{-5pt}
Kentaro Inui${}^{\,\ddagger, \dagger}$
\quad\hspace{-5pt}
\\[5pt]
${}^{\ddagger}$ Tohoku University \quad 
${}^{\dagger}$ RIKEN Center for Advanced Intelligence Project \quad
${}^{\star}$ Stony Brook University\\
{\tt \{\hspace{0em}keshav.singh29,\hspace{0em}inui\}@ecei.tohoku.ac.jp}\\
{\tt paul.reisert@riken.jp}\\
{\tt naoya.inoue.lab@gmail.com}
}

\begin{document}
\maketitle
\begin{abstract}
Explicating implicit reasoning (i.e. \textit{warrants}) in arguments is a long-standing challenge for natural language understanding systems. 
While recent approaches have focused on explicating warrants via crowdsourcing or expert annotations, the quality of warrants has been questionable due to the extreme complexity and subjectivity of the task.
In this paper, we tackle the complex task of warrant explication and devise various methodologies for collecting warrants.
We conduct an extensive study with trained experts to evaluate the resulting warrants of each methodology and find that our methodologies allow for high-quality warrants to be collected.
We construct a preliminary dataset of 6,000 warrants annotated over 600 arguments for 3 debatable topics. 
To facilitate research in related downstream tasks, we release our guidelines and preliminary dataset.
\end{abstract}

\section{Introduction}
In daily conversations, humans create many arguments consisting of implicit, unstated reasoning. 
Consider the following example consisting of two argumentative components: \textit{Claim} (i.e., declarative statement) and its \textit{Premise} (i.e., supporting statement):

{\enumsentence{\textbf{Claim}: We should introduce compulsory voting.\\
\textbf{Premise}: It increases voter turnout.} 
\label{ex:1}}

In Example~\ref{ex:1}, it is easy for humans to understand the reason why the premise supports the claim. For example, one might assume that it implies that high voter turnout, resulting from introducing compulsory voting, is good for a fair representation of society.
However, for machines, explicating such implicit reasoning, henceforth \textit{warrants}~\cite{toulmin1958use}, is a challenging, but exciting task.

Re-constructing warrants in arguments can play an important role in many argument-related downstream applications, such as argumentative writing and comprehension support~\cite{qin2010analysis, von2019improve}, improving students critical thinking~\cite{hillocks2010ej}, and debate systems~\cite{weber2008learning}.
For some of these applications, assessing identification of warrants is an important step in developing a more complex pipeline for overall argument analysis~\cite{becker2020explaining}.
Furthermore, warrants can be leveraged to automatically give constructive feedback to users to assist them in the aforementioned argumentative applications. 


While the identification of main argumentative components such as claim and premise have been well explored~\cite{habernal2017argumentation, ein2020corpus}, research focusing on warrant explication has been lacking due to its extreme complexity. 
\newcite{habernal-etal-2018-semeval} and~\newcite{boltuzic-snajder-2016-fill} both demonstrated that simple warrant annotations can be done through crowdsourcing, though the authors acknowledge the high complexity due to subjectivity for reconstructing warrants. 
They attribute the difficulty to the variety of reasoning patterns possible for framing warrant and multiplicity of warrants between argumentative components.
However, to the best of our knowledge, no work has yet to solve the aforementioned challenges.



To tackle the challenge encountered in collecting warrants, in this paper we address the following research question: \textit{1. Can we devise a methodology that makes it easier to construct high quality warrants?}.
Our main hypothesis is that since there are multiple reasoning patterns in natural language~\cite{hitchcock2003toulmin, kock2006multiple}, it must be possible to narrow down the warrants into specific keyword-based patterns, where keywords are utilized from the original argument (i.e., claim and premise), which follows the formal definition of warrants in that they serve as a linking chain between the contents of claim and its premise~\cite{freeman1992relevance, verheij2005evaluating}.

In this work, we conduct an extensive annotation study with trained experts and crowd workers to compare three different methodologies for collecting warrants, namely: \textit{Natural language warrants, User-defined Keyword-based warrants} and \textit{Pre-defined Keyword-based warrants}.
Firstly, through each methodology, we collect multiple warrants on top of argumentative texts from IBM-Rank-30k dataset~\cite{gretz2019large}, a dataset of topic and claim pairs already annotated with quality scores. 
We analyze the collected warrants and work closely with experts to adapt the annotation task, iterating over how best to approach the novel domains and simplify the annotation guidelines for crowdsourcing to be suitable for both expert and non-expert annotators. 
Secondly, based on our extensive qualitative analysis results on the methodologies, we collect 6,000 warrants annotated for 600 arguments from over 6 diverse topics via \textit{User-defined Keyword-based warrants}. 
We publicly release our guidelines and preliminary dataset to facilitate research in automatic warrant generation.\footnote{\url{https://github.com/keshav2995/6000_warrants}}







\section{Related Work}

Implicit reasonings, commonly referred to as \textit{warrants}~\cite{toulmin1958use} or \textit{enthymemes}~\cite{walton2008argumentation}, have long been studied to understand the grounds on which a premise lends support to the topic~\cite{freeman1992relevance}.
In other words, a warrant, when made explicit, clearly shows the principle upon which the argument rests~\cite{pineau2013abuses}.
Identification of such warrants has been shown to aid in the argument comprehension process~\cite{hitchcock2006arguing, becker2017enriching} and in educational research for helping students to make better arguments and improve their critical thinking skills~\cite{erduran2004tapping,von2019improve}.

Numerous computational approaches have been made towards solving the task of warrant explication in arguments.
In an initial attempt,~\newcite{feng-hirst-2011-classifying} identified argumentation schemes~\cite{walton2008argumentation} as a means for warrant reconstruction, but they do not approach the task due to absence of training datasets. 
\newcite{boltuzic-snajder-2016-fill} made a preliminary attempt at crowdsourcing warrants to fill the reasoning gap between a claim and premise. 
They let each worker write multiple warrants to fill the reasoning gap between a claim and premise pair without any specific guidelines. 
However, they concluded that the collected warrants were highly variable in wordings and amount needed to fill the gap.
In contrast, our crowdsourcing methodologies provide specific guidelines with numerous examples and restrict workers to write only the most relevant warrant.

\newcite{becker2017enriching} hire expert annotators to fill missing knowledge pieces between different argument units (counter-argument, topic, premise, rebuttal, etc) in the argmicrotext corpus~\cite{peldszus2015annotated}.
However, their approach relies on the availability of experts which is an expensive and time-consuming process for collecting warrants on a larger dataset.
Instead, we perform crowdsourcing with non-experts and show that the quality and cost of warrants collected can be high and inexpensive. 
Recently,~\newcite{habernal-etal-2018-semeval} proposed a step by step methodology to crowdsource warrants from non-experts, but their error analysis pointed at inherent difficulty in distinguishing patterns between incorrect warrants. 
On the contrary, in our work, we create various methodologies which can make it easier to analyse good and bad warrants for a given argument.

\section{Dataset Desiderata}
Towards creating a preliminary warrant annotated dataset which can be potentially useful in downstream argumentative applications, we require it to fulfill at least the following criteria: i) cover multiple topics, ii), span over diverse premises, and iii) consists of high-quality warrants.
In addition, the dataset creation methodology must be cost effective without compromising data quality and easier for experts as well as non-experts to understandably write quality warrants.
As mentioned a priori, a dataset with multiple topics is desirable; however, collecting warrants across multiple topics can be both difficult and time consuming.
Thus, we start with small number of topics and define a simple metric to filter a handful of topics (specifically, 3) found in a large, well-known argumentation dataset of diverse premises (\S~\ref{sec:sourceargs}).

\subsection{Source Data (IBM-Rank-30K)}

\begin{table}[t]
    \small
    
     \begin{tabular}{ l r r }
      \toprule
         Topic $(t)$: \textbf{We should} & $(dp_t)$ & \# Premise \\ 
         \midrule
         Abolish the Olympic Games & 2.68 & 204  \\
         Ban missionary work & 2.57 & 218\\
         Ban the Church of Scientology & 2.50 & 195 \\
         Abolish safe spaces & 2.34 & 194 \\
         Oppose collectivism & 2.27 & 225 \\
         \midrule
         Fight urbanization & 1.45 & 202\\
         \textbf{Abolish zoos} & \textbf{1.41} & 198\\
         \textbf{Introduce compulsory voting} & \textbf{1.37} & 205\\
         \textbf{Ban whaling} & \textbf{1.36} & 217\\
         Mandate use of public defenders & 1.25 & 218\\
         \midrule
         Overall & 2.92 & 300 \\
    \bottomrule
    \end{tabular}%
    \caption{Topics in \emph{IBM-Rank-30K} with top-5 and bottom-5 premise diversity growth rate$(dp_t)$. The overall premise diversity of IBM data as shown above is calculated for random sample of 300 premises across various topics.} 
    \label{table:diversity}
\end{table}

\label{sec:sourceargs}
We utilize \emph{IBM-Rank-30K} dataset~\cite{gretz2019large}, which contains supporting and opposing stance arguments on 71 common controversial topics.
Several factors motivated our choice of utilizing this dataset:
(1) Arguments were collected actively from crowdworkers with strict quality control measures as opposed to being extracted from targeted audiences such as debate portals.
This represents a vast majority of all the possible arguments that can be made for a given topic.
(2) Argument structure is not tree-like (i.e., a premise supporting another premise, etc.) but rather flat-list like (i.e., a premise offers direct support to the topic). 
We believe this will be useful in the pedagogical context, where students may make a declarative statement and generally provide few premises.
(3) The arguments have already been annotated with point-wise quality for both supporting and opposing stance. 
Our extension on a small subset of it could serve the needs of similar qualitative research in future.\footnote{We only utilize arguments in  \emph{IBM-Rank-30K dataset} with supporting stance (i.e., the premise supports the topic) since warrants only exist for arguments which lend support to topic \cite{toulmin1958use}.}

To select the most promising topics for testing our crowdsourcing methodologies, we define a simple metric. 
Specifically, for each topic$(t)$, we estimate the Growth Rate (GR) i.e. slope of diversity of premises$(dp_t)$ as a function of its vocabulary size (V):

\begin{equation}
    dp_t = \frac{(|V(p_t)| - |V(p_t^{50\%})|)}{(|p_t| - |p_t^{50\%}|)},
\end{equation}

where $p_t$ is a set of premises associated with $t$, $p_t^{50\%}$ is a 50\% random sample of $p_t$, and $V(p)$ is a set of unique words (i.e. vocabulary size) of $p$ obtained after tokenization and lemmatization.
The final growth rate of premises ($dp_t$) for each topic is calculated after averaging over multiple random resampling runs.


As shown in Table~\ref{table:diversity}, the variety of premises for topics with a lower $dp_t$ depicts higher number of overlapping premises while topics with a higher $dp_t$ might comprise of more variety of debatable premises. 
It also indicates that the growth rate of diversity strongly depends on the topic i.e. certain topics contain more diverse keywords/knowledge pieces as compared to others.
To proceed with our preliminary warrant crowdsourcing and methodology analysis, we choose 3 topics that fall in the lower band of $dp_t$ value. 
Our choice for the above topics is based on the assumption that if high quality warrants can be collected for relatively low diverse premises then it will be a strong evidence to focus on more diverse premise in future work.


\section{Annotation Study}
In this section, we detail how we developed and designed our annotation task to allow for an efficient, reliable collection of warrants with crowdsourcing. 
In particular, we extensively analyze 3 different warrant crowdsourcing methodologies and highlight the benefits of utilizing the best one through a comparative study.

The aim of our annotation is to reveal warrants that connect premises in argumentative texts to its claim. 
Explicating warrants, i.e., making the implicit explicit, has been inherently difficult due to varying structural or lexical phrasing chosen by annotators, difference in intuition, and a variety of reasoning patterns in which warrants can be framed.
To overcome these challenges, we attempt to conduct a deeper analysis on warrants from a theoretical perspective of reasoning and explore different ways to collect these warrants. 

\subsection{Preliminary Measures}
We employ several preliminary measures to ensure the quality of collected annotations for all of our crowdsourcing tasks. 
We choose Amazon Mechanical Turk (AMT)\footnote{\url{www.mturk.com}} as our crowdsourcing platform due to its success in previous argumentation mining tasks~\cite{habernal-etal-2018-semeval}. As an initial step, we only allowed workers who had $\geq$ 99\% acceptance rate, $\geq$ 5,000 approved Human Intelligence Tasks (HITs) and 100\% score on our custom Reasoning Qualification Task (RQT) to work on the tasks.

Additionally, to address any ethical issues~\cite{adda2011crowdsourcing} raised by our task, we actively monitored multiple pilot tests to ensure workers were satisfied with our task. Simultaneously, we corresponded directly to workers that had questions/comments on our task.
Crowdworkers are paid in accordance with the minimum wage calculated by conducting many trials and based on average work-time. 

\subsection{Warrant Collection Methodologies}

\begin{figure}[t]
    \centering
    \includegraphics[width=\linewidth]{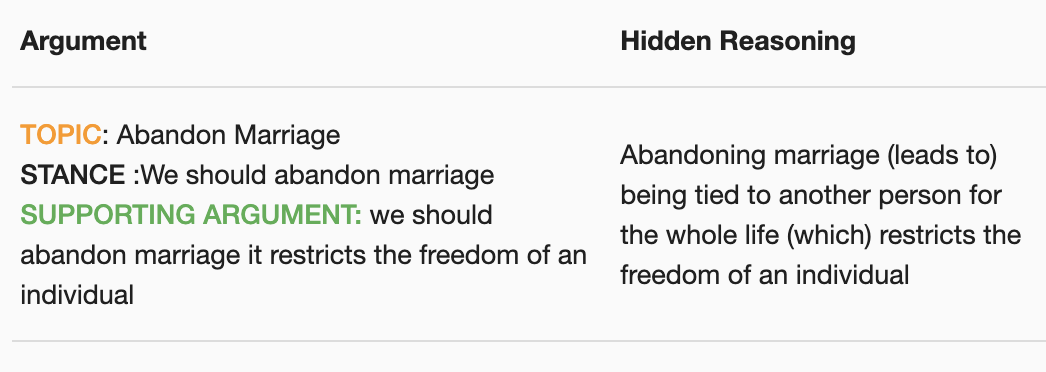}
    \caption{An example of a warrant collected for a claim and premise pair during our User-defined Keyword-based warrant collection trial. We note that claim and premise were shown as 'stance' and 'supporting argument' to avoid using complex terminology during our experiments.}
    \label{fig:my_label}
\end{figure}

Our choice for the warrant collection methods is based on the formal definition of warrants i.e. warrants are the inference link that fills the reasoning gap between claim and premise~\cite{toulmin1958use}. 
Hence, we consider 3 methodologies: \textit{natural language warrants}, \textit{pre-defined keyword-based warrants}, and \textit{user-defined keyword-based warrants}.
Bonus is paid to workers based on general consensus rather than sole judgement which may help remove bias towards and between workers.

\paragraph{Natural Language Warrants}
For collecting warrants in free-form natural sentences, we follow guidelines similar to~\newcite{habernal-etal-2018-semeval}.    
Prior  to  the  main  annotation  task, we  filtered crowdworkers via our custom Reasoning Qualification Test (RQT).
Selected annotators were presented with a topic and single premise and asked to come up with a warrant that indicates why the premise supports the claim. 

\paragraph{User-defined Keyword-based Warrants}

As an alternative to natural language sentences which can be variable in structure due to myriad amount of reasoning patterns~\cite{keith2008toulmin}, we follow a simpler way to restrict the crowdworkers to a specific reasoning pattern. 
To do this, we follow \newcite{reisert-etal-2018-feasible}'s argumentation slot-filling templates, where each template encodes an argumentative relation between components. Such templates enable annotators to focus on important keywords in components which are relevant to the missing/implicit information linking a claim and a premise. 

Following \newcite{reisert-etal-2018-feasible}'s annotation study in which expert annotators freely chose slot-fillers, we let the crowdworker freely choose the slot-fillers from both the claim and the premise and additionally fill in the missing/implicit reasoning between them.
For example, Fig.~\ref{fig:my_label} shows a case in which the warrant connects the ontological elements of both claim and premise, and the claim is a consequence resulting from the premise (i.e., argument from consequence~\cite{walton2008argumentation}).

\paragraph{Pre-defined Keyword-based Warrants}
\begin{table*}[t]
    \centering
    \adjustbox{max width=\linewidth}{%
    \begin{tabular}{lrrrrrr}
    \toprule
    \multirow{1}{*}{Methodology}&
    \multicolumn{2}{c}{Natural Language} & \multicolumn{2}{c}{User-defined} & \multicolumn{2}{c}{Pre-defined}  \\\cmidrule(rr){2-3}\cmidrule(rr){4-5}\cmidrule(rr){6-7}

    & $\alpha$ & Avg. & $\alpha$ & Avg. & $\alpha$ & Avg.\\\midrule
    Abolish zoos & 0.64 & 1.55 & 0.67 & 1.60 & 0.62 & 1.57\\
    Intro. compulsory voting & 0.45& 1.50 & 0.51 & 1.55 & 0.53 & 1.47 \\
    Ban whaling & 0.63 & 1.37 & 0.61 & 1.61 & 0.58 & 1.59\\
    \midrule
    Overall & \textbf{0.57} & 1.46 & 0.56 & \textbf{1.59} & 0.53 & 1.55 \\
    \bottomrule
    \end{tabular}}
    \caption{Comparison between the different warrant crowdsourcing methodologies. 
    Krippendorff's $\alpha$ and Average scores given by two expert annotators (Avg.) on a scale of (0-2) indicate that user-defined and pre-defined methodologies result in overall higher quality warrants as compared to natural language warrants.}
    \label{tab:method_results}
\end{table*}
To follow-up on \newcite{reisert-etal-2018-feasible}'s annotation method, we also would like to determine the quality of warrants when the slot-fillers are pre-defined. Such a method allows for annotators to only write the implicit information between pre-defined keywords from the claim and premise.
For choosing keywords, we employ spaCy~\cite{spacy} and automatically choose the longest verb/noun phrases from the claim and premise.

\section{Methodology Comparison and Results}
\begin{table}[t]
    \begin{tabularx}{\columnwidth}{l|X}
        \textbf{Score} & \textbf{Explanation} \\
        \toprule
         \textbf{0} & Warrant is unrelated to the topic and its premise. \\
         \textbf{1} & Warrant is related to the topic and premise but does not make the relationship between them easy to understand and/or strengthen the argument. In addition, the warrant may overlap or be a paraphrase of the premise. \\
         \textbf{2} & The relationship between the topic and premise is easier to understand and/or strengthened because of the warrant.\\
         \bottomrule
    \end{tabularx}
    \caption{Guidelines used by our expert annotators for scoring the quality of warrants on a scale of 0-2.}
    \label{tab:quality_guidelines}
\end{table}

Towards determining which methodology results in the highest quality warrants, we first collect warrants for each methodology using crowdsourcing. For each methodology, we annotate 40 arguments covering 3 different topics with 5 crowdworkers per argument. First, crowdworkers decided whether a warrant can be constructed between the given claim and premise. If so, they were then asked to write the warrant in the correct format.

\subsection{Filtering}
For each methodology, we can collect, at most, 200 warrants, if a worker first identifies that hidden reasoning is necessary to link the claim and its premise.
However, after initial filtering we discovered that workers wrote 65, 86 and 80 warrants each respectively out of a possible 200 warrants for the arguments given to them for \textit{Natural language warrant, User-defined keyword-based} and \textit{Pre-defined keyword-based warrant} methodologies. 
We utilize these 231 collected warrants for further analysis.

\subsection{Results}
To analyze the quality of the collected warrants, two expert annotators were asked to judge the quality of the warrants on a scale of 0-2 according to the guidelines as shown in Table~\ref{tab:quality_guidelines}. 
Both annotators judged 50 warrants randomly chosen from the pool of collected warrants from each methodology. 
The Inter-Annotator Agreement (IAA) score between the experts was measured as Krippendorff's alpha ($\alpha$)~\cite{krippendorff2011computing}. We also measure the combined average score given to the warrants for each methodology to measure the quality of warrants as shown in Table~\ref{tab:method_results}.

\paragraph{Natural language} Expert annotators achieved an overall agreement score of $\alpha$ = 0.57, indicating a good agreement. The average score given to natural language warrants by both annotators was 1.46. 
Overall, the experts have the highest agreement for warrants collected via this method, but the average score for warrant quality was lowest in this method.

\paragraph{User-defined} Expert annotators achieved an overall agreement score of $\alpha$ = 0.56. The average score given to natural language warrants by both annotators was 1.59 which is the highest overall. 
This suggests that warrants collected via this method might result in higher overall quality of warrants. 

\paragraph{Pre-defined} Expert annotators achieved an overall agreement score of $\alpha$ = 0.53. The average score given to natural language warrants by both annotators was 1.55. However, since the difference between pre-defined and user-defined warrant collection results is not significant in terms of the Average scores, it may require additional analysis to compare the two methods.

The overall results are summarized above in Table~\ref{tab:method_results} and suggest that the difficulty of the task is highly dependent on the method of warrant collection and also on the topic to some extent. 
Between the 3 methodologies, the average quality scores were highest with user-defined methodology and lowest with natural language.

\subsection{Qualitative Analysis}

To further analyze the quality of warrants and the quality of the entire crowdsourcing process, we further analyze a sample of the warrants collected via each methodology and annotated by experts. 
\begin{figure*}[t]
    \centering
    \includegraphics[width=\linewidth]{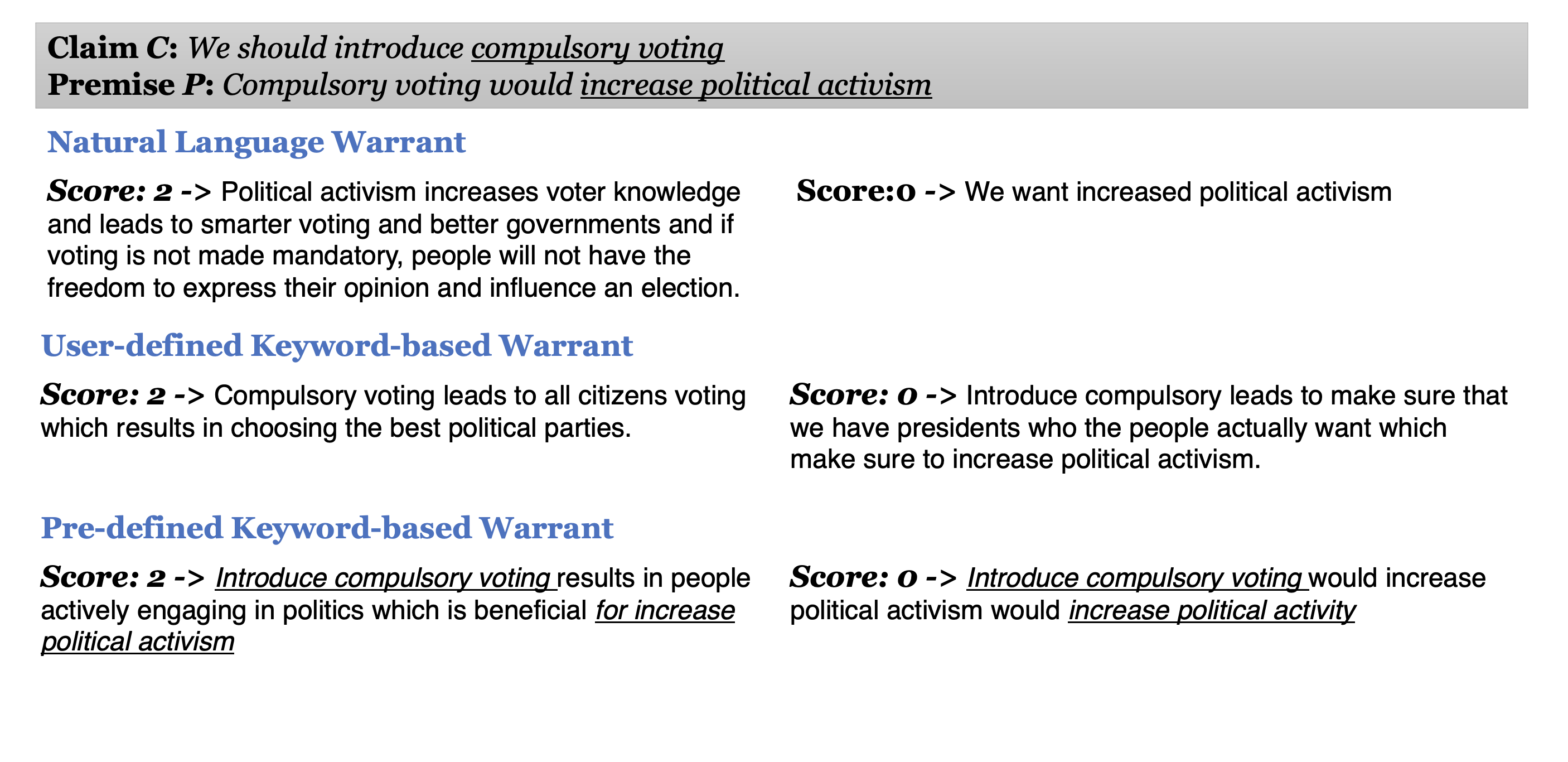}
    \caption{Sample from the warrants collected for our methodology comparison.
    These warrants were scored the same from both expert annotators.
    The underlined text denotes keywords from \textbf{Premise} and \textbf{Claim} used in pre-defined keyword-based warrant methods. 
    }
    \label{fig:example_warrants}
\end{figure*}

As shown in Fig~\ref{fig:example_warrants}, the warrants shown each have some kind of keyword overlap with the claim/premise. 
However, the difference in scoring warrants might be attributed to different complexities of the way in which warrants are framed.
Specifically, even though the warrants encode claim-premise information, the quality of the warrant can essentially be bad.
This shows that identifying the correct warrant can still be challenging with automated techniques which extract lexical and word-level features.

While keyword-based methods restrict most warrants to a single sentence, the natural language warrants often consist of shorter, multiple sentences, even with character restrictions.
Overall, we found 23\% of natural language warrants were compound sentences.
Also, such warrants were often found to be paraphrased from the information in the premise or workers simply used a previous premise in place of the warrant. 
Although this was mostly seen in natural language warrants, we found a few instances of such type in keyword-based methods also.
This might suggest that, although keyword-based methods reduce bad quality warrants in-terms of paraphrasing and incomprehensible warrants, there is still room for improvement.

\section{Preliminary large-scale corpus}
Based on our finding that user-defined keyword-based warrant method comparatively results in good warrants, we follow this method to collect a total of 6,000 warrants across 3 topics, annotated for 600 claim-premise pairs. All warrants were limited to have a length between 60 and 200 characters. 
Since this is an ongoing work, we plan to make further analysis on our preliminary dataset in future.






\section{Conclusion}

In this work, we tackle the difficult task of explicating warrants in arguments. Due to the complexity of collecting warrants using natural language approaches, we conduct an extensive analysis and annotation studies for determining an appropriate methodology for collecting warrants and determine a user-defined keyword approach results in the highest quality warrants. Based on our finding, we construct a preliminary corpus of 6,000 warrants and make the warrants and guidelines publicly available.\footnote{\url{https://github.com/keshav2995/6000_warrants}} 

In our future work, we will annotate our corpus with quality dimensions shown from previous argumentative works~\cite{wachsmuth2017computational}.
Simultaneously, we will collect warrants for premises attacking the topic, as our main focus of this work was on premises supporting the original topic. We will also test the usefulness of our model as constructive feedback in a pedagogical setting, such as deploying our system in schools in which students debate and/or write essays on controversial topics.
Finally, we will also focus on argumentation in tree-like hierarchical structures, such as a premise supporting/attacking another premise, which can be discovered in everyday, real-world arguments.


\section*{Acknowledgements}
This work was partially supported by JST CREST Grant Number JPMJCR20D2 and a project, J200001946, subsidized by the New Energy and Industrial Technology Development Organization (NEDO).

\bibliography{emnlp2021}
\bibliographystyle{acl_natbib}

\appendix



\end{document}